\documentclass[conference,a4paper]{IEEEtran}
\ifCLASSINFOpdf
\else
\fi

\usepackage{algorithm}
\usepackage{algpseudocode}
\usepackage{graphicx}
\usepackage{amsmath}

\begin{document}
%
\title{A Tool-Augmented, GPT-4 Chatbot for Real-Time Repository Data Analysis}

\author{\IEEEauthorblockN{Muhammad Jawad Chowdhury}
\IEEEauthorblockA{Computer Science and Engineering\\
Islamic University of Technology\\
Dhaka, Bangladesh\\
Email: jawad@iut-dhaka.edu}
\and
\IEEEauthorblockN{Md. Sakib Khan}
\IEEEauthorblockA{Computer Science and Engineering\\
University of Dhaka\\
Dhaka, Bangladesh\\
Email: sakibkhan111296@gmail.com}
}


%


\maketitle

\begin{abstract}
Software repositories contain vast amounts of data on code contributions, bug reports, and project activities, yet this information remains challenging for non-technical stakeholders and developers to access due to limited expertise in querying repositories. To address this, we introduce a novel chatbot architecture leveraging OpenAI’s GPT-4 model for automated extraction and analysis of repository data. In contrast, our architecture takes a structured path first by parsing the user’s query to extract relevant parameters, then selecting the correct tool to employ based on that analysis, and finally invoking the GPT-4 model to create a highly detailed response. In contrast to previous work based on multi-component systems with embedding models and document retrievers, our architecture inverts the process by relying on prompt engineering and tool selection to fit with the query intent. To validate our approach, we conducted experiments on various question types, including (Issues, Pull Requests, Commits, Compound Questions, and General Repository Information) evaluating our target prompts’ ability to improve the accuracy of responses from the model. Beyond demonstrating the utility of this architecture to a diverse set of users, our findings suggest that this architecture can make repository data more accessible to technical and non-technical audiences through the production of actionable insights.


\end{abstract}


%
\IEEEpeerreviewmaketitle

\section{\textbf{Introduction}}
In the context of open-source software, platforms such as GitHub act as an essential source and database of large amounts of data created by multiple contributors \cite{kalliamvakou2015open}\cite{borges2016predicting}\cite{borges2016understanding}. This data should be analyzed to gain more insights that would improve the quality of software and also increase our understanding of the software development process \cite{gousios2008measuring}\cite{chatziasimidis2015data}\cite{hu2016influence}. However, analyzing repository data is challenging because repositories contain many interconnected elements, including commits, pull requests, issues, and contributor activities. Extracting useful insights from these elements can be time-consuming and requires technical expertise. \cite{abedu2024llm}\cite{luo2024repoagent}\cite{proma2024visual}.

Also, the involvement of non-technical stakeholders in the analysis of the repository has numerous benefits \cite{capiluppi2012exploring} such as better collaboration \cite{gousios2008measuring}, clearer communication, and better decision-making \cite{maturana2004being}\cite{mcmanus2004stakeholder}. Yet existing barriers prevent these stakeholders from accessing or understanding repository data. The most crucial property of iteration mechanisms is to generate summary reports \cite{khleel2020mining} on demand so that non-technical users can monitor project progress, identify risks, and make data-driven decisions without needing to rely solely on developers \cite{chaturvedi2013tools}\cite{de2016systematic}.



Large Language Models (LLMs) on the other hand, have rapidly evolved to become central to a wide array of applications, ranging from natural language processing tasks like translation and summarization to creative writing \cite{brown2020language}, content generation \cite{radford2019language}, and even healthcare \cite{nerella2023transformers} and customer support \cite{mctear2022conversational}. These versatile models leverage vast amounts of training data to produce human-like responses, making them valuable tools across industries \cite{hou2023large}\cite{ozkaya2023application}. 
 Thus, researchers and developers are beginning to harness LLMs for in-depth software design analysis, highlighting LLMs' growing role in advancing the capabilities and efficiencies within software development processes \cite{fan2023large}\cite{hu2024leveraging}\cite{ross2023programmer}\cite{xiao2023empirical}. In line with this, our research introduces a chatbot solution for GitHub repositories \cite{abedu2024llm}\cite{abdellatif2020msrbot}, designed to simplify the process of querying and analyzing repository data, thereby enhancing the software development process for both technical and non-technical users \cite{xu2017answerbot}\cite{beschastnikh2017accelerating}\cite{tian2017apibot}\cite{wessel2018power}.

The key contributions of our chatbot are as follows:
\begin{enumerate}

    \item We propose a tool-augmented architecture that parses natural-language repository queries, extracts relevant parameters, selects appropriate tools, and retrieves real-time data through the GitHub REST API.
    
    \item We construct an evaluation dataset of 80 repository-related questions across pull requests, commits, issues, compound queries, and general repository information, including out-of-scope questions for testing abstention behavior.
\end {enumerate}    

\section{\textbf{Related Work}}

\subsection{LLMs in Software Engineering}
 Recently, LLMs have been used in several software engineering contexts such as code generation \cite{lin2024llm} \cite{ouyang2023llm}, program repair \cite{bouzenia2024repairagent} \cite{yang2024cref}, and code summarization \cite{ahmed2022few}\cite {ahmed2024automatic}. For example, Fan et al. \cite{fan2023automated} used Automated Program Repair (APR) techniques to demonstrate that LLMs, e.g., Codex, are better at generating correct fixes than traditional tools when applied to programming tasks. Likewise, Wang et al. \cite{wang2023codet5+} present 'CodeT5+', an LLM that supports a variety of downstream code tasks.
 Other recent works have taken LLM applications beyond research to the area of software engineering. In a paper by Feng et al. \cite{feng2020codebert}, 'CodeBERT' an LLM specifically trained to increase semantic code understanding for tasks like function extraction and code-to-text generation was developed. Ahmed et al. \cite{ahmed2024automatic} explored the ability of prompt engineering for the LLM to improve accuracy in code summarization tasks and found that a carefully crafted prompt can substantially improve automated documentation generation.

\subsection{Software Engineering Chatbots}

Abdellatif et al. \cite{abdellatif2020msrbot} proposed MSRBot, a chatbot that enables developers to submit natural language queries to retrieve answers from the repository data. They found that MSRBot helped participants complete the tasks more quickly and accurately than manual methods. Xu et al. \cite{xu2017answerbot} also proposed AnswerBot, a chatbot to summarize stack overflow multi-answer posts to help developers get concise, short, and relevant answers. Most recently Yu et al. \cite{yu2022code} presented CodeMaster, a chatbot that answers code-related questions using CodeT5, showcasing the appearance of LLMs in chatbots that answer complex queries for developers. Other work has also attempted to enable chatbots to answer developer questions through information retrieval. Gottipati et al. \cite{gottipati2011finding} have developed a semantic search engine that retrieves answers from threads related to software, while Bradley et al.\cite{bradley2018context} have built a conversational assistant relying upon the Amazon Alexa platform to automate developer tasks such as creating pull requests.

\section{\textbf{Approach}}

\begin{figure*}[t]
  \centering 
  \includegraphics[width=\textwidth]{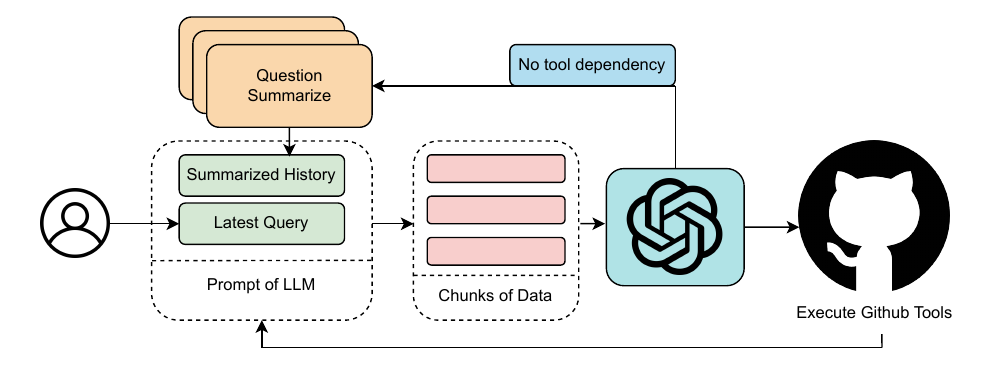}
  \caption{Architecture of the repository interaction – \textbf{Prompt of LLM:} It consists of Summarized history or response we get from GPT-4 with the latest query and System message \textbf{Data Chunk:} The data or query is chunked based on the limit \textbf{Response Generation:} Necessary tools and parameters are recognized through iterative process.}
  \label{fig:framework}
\end{figure*}

\subsection {\textup{\textbf {User Interface:}} }
The interaction begins with a user interface, where the user submits a single input \( Q \) consisting of a GitHub repository URL \( U_{\text{repo}} \) and an accompanying \textbf{query} related to the repository. Once submitted, the system constructs an input \( I \) to be processed by the model, composed of three key components: \textbf{system messages} \( M_{\text{sys}} \), \textbf{summarized messages} \( M_{\text{sum}} \), and the user’s \textbf{latest query} \( Q_{\text{user}} \).

1. \( M_{\text{sys}} \): Provides context by defining the chatbot’s responsibilities and response format.


2. \( M_{\text{sum}} \): A history of previous user queries \( Q_{1}, Q_{2}, \dots, Q_{n-1} \) (excluding responses) is appended to the latest query to maintain context in multi-turn conversations, ensuring continuity without exceeding the model’s token limit.

3. \( Q_{\text{user}} \): The most recent query submitted by the user, which is combined with \( M_{\text{sys}} \) and \( M_{\text{sum}} \) to form the final input \( I \) for the chatbot's response generation.

Thus, the complete input \( I \) for processing by the chatbot is:
\[
I = \{ M_{\text{sys}}, M_{\text{sum}}, Q_{\text{user}} \}
\]
The chatbot uses \( I \) to generate responses that address the query within the structured context provided by \( M_{\text{sys}} \) and \( M_{\text{sum}} \).




\subsection {\textup{\textbf {System Prompt:}} }

The system prompt is a pre-defined instruction given to the chatbot at the beginning of the interaction. It provides a detailed description of the chatbot's role, its capabilities, and the context in which it operates. The format of the prompt is as follows:

 \textit  {"You are a GitHub Repository Analysis Agent named 'Github-chatbot.' Answer the following questions as best you can on GitHub Repository: \{self.github\_url\}. You have access to the following tools: \{tools\_details\}}. 


The system prompt consists of several components:

\begin{itemize}
    \item \textbf{Agent Identification}: The prompt begins by clearly defining the chatbot's role as a "GitHub Repository Analysis Agent" named \textbf{'Github-chatbot'}. This helps establish the chatbot's identity and purpose, ensuring the user understands that the chatbot is specifically designed for GitHub repository analysis.
    
    \item \textbf{Repository URL Placeholder}: The placeholder \{ self.github\_url \} is used to dynamically insert the URL of the specific GitHub repository being analyzed. This ensures that the chatbot knows which repository to focus on when answering user queries.
    
    \item \textbf{Tool Access Information}: The prompt specifies the tools available to the chatbot for answering queries. The placeholder \{ tools\_details \} is dynamically populated with a list of tools that the chatbot has access to. These tools could include:
    \begin{itemize}
        \item \textbf{Repository Report Tool}: Provides comprehensive insights into the repository's structure, metadata, and contributions.
        \item \textbf{Commits/Issues/PR Report Tool}: Focuses on retrieving detailed information about commits, issues, and pull requests within the repository.
    \end{itemize}
\end{itemize}

\subsection {\textup{\textbf{Query Processing:}}}


Given an input query \( Q \), the system first parses and classifies it to identify key components. The analysis focuses on extracting essential elements and selecting the appropriate action based on predefined keywords and structural rules.

\begin{itemize}
    \item \textbf{Data Classification}: Using a keyword-matching function \( f: Q \to C \), where \( C \) represents data categories, the system directs \( Q \) to the relevant GitHub repository data stream. Primary categories include:
    \begin{itemize}
        \item \textbf{Issues},
        \item \textbf{Pull Requests}, and
        \item \textbf{Commits}.
    \end{itemize}
    
    \item \textbf{Filter Application}: User-specified filters \( F = \{ f_1, f_2, \dots, f_n \} \) are applied to refine \( Q \), where \( f_i \) may denote criteria such as issue state (open/closed), author, date range, or result limit. This ensures that the subset \( Q_F \subset Q \) aligns closely with user requirements, yielding targeted and relevant results.
\end{itemize}

\subsection {\textup{\textbf{Tool Selection and API Interaction:}}}

Based on the result of the query analysis, the system determines whether specialized tools are required to retrieve or transform the requested data. A \emph{tool} in this context refers to a predefined set of operations designed to interact with the repository and extract specific information as per the user's query.

\begin{itemize}
    \item \textbf{Repository Report Tool}: This tool is used for generating comprehensive reports on the repository's metadata and overall structure.
    \item \textbf{Commits/Issues/PR Report Tool}: This tool focuses on handling specific data points related to Commits, Issues, and Pull Requests (PRs).
\end{itemize}

Once the appropriate tool is selected, the system formulates an API request using the input parameters (repository URL, endpoint, scope, and filters). The interaction is handled via GitHub's REST API, which fetches the required data.



\subsection {\textup{\textbf{Response Generation}}}

The response generation process adapts based on whether a tool is selected or not, following an algorithm:

\begin{algorithm}
\caption{Response Generation Process}
\begin{algorithmic}
    \State \textbf{Input:} Query $Q$, Interaction History $H$, Tool Availability Flag $T$
    \If{$T = \text{False}$}
        \State \textbf{No Tool Selected:} Generate response using internal knowledge.
        \State $Response \gets \text{GPT}(Q, H)$
    \Else
        \State \textbf{Tool Selected:} Generate response with tool-based data.
        \For{each retrieved data chunk}
            \State Process the chunk and integrate it into the response.
        \EndFor
    \EndIf
    \State \textbf{Return} $Response$
\end{algorithmic}
\end{algorithm}

This conditional framework provides flexibility for integrating future tools and algorithms as needed, allowing modular expansion of response generation capabilities.



\section{\textbf{Dataset}}

\begin{table}[ht]
\centering
\begin{tabular}{|c|p{5cm}|} 
\hline
\multicolumn{1}{|c|}{\textbf{Category}} & \multicolumn{1}{c|}{\textbf{Question}}  \\
\hline
Pull Request & How many open pull requests are there currently? \\
\hline
Commit & What is the latest commit message? \\
\hline
Compound & Show me the number of commits and the number of pull requests. \\
\hline
Issues & How many open issues are there?\\
\hline
General Info & Which developers contributed the most code in 2024?\\
\hline
\end{tabular}
\caption{The table consists of all the categories of questions we have in the dataset.}
\label{table:table_question_category}
\end{table}

\begin{figure}[ht]
  \centering
  \includegraphics[width=0.8\columnwidth]{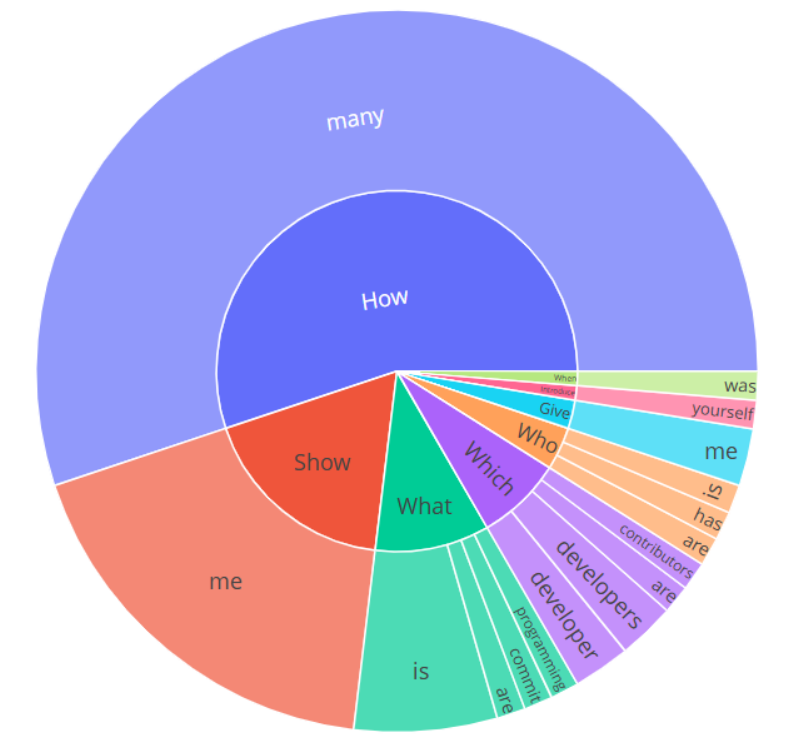}
  \caption{Sunburst Distribution of the first two words of the dataset.}
  \label{fig:sunburst}
\end{figure}

We created a custom dataset of 80 questions to evaluate the proposed approach across a range of repository-related query types. The questions were derived from 15 exemplar queries presented by Abdellatif et al. \cite{abdellatif2020msrbot}, which describe intents and entities used in repository analysis. Drawing on those exemplars, we prompted an AI model to generate 80 semantically relevant variations, preserving the original intents while diversifying entities and phrasings. Figure \ref{fig:sunburst} illustrates the variety of questions by showing how each question begins. These questions were divided into five distinct categories: \textbf{Pull Requests}, \textbf{Commits}, \textbf{Compound Questions}, \textbf{Issues} and \textbf{General Information}. Table \ref{table:table_question_category} provides examples for each category. Pull Requests assess the model's ability to understand and interact with GitHub’s Pull Request workflows. Questions in the Commits category evaluate the model's understanding of commit history and version control. Compound Questions are designed to test the model’s capacity to handle multi-faceted queries that may involve combining information from multiple sources within the repository. Issues test the model’s ability to interpret and extract relevant information about open or closed issues. Lastly, General Information includes questions about basic repository details and general metadata, which test the model’s ability to provide high-level information about the repository.

It is designed to assess model performance and determine if the models select appropriate metrics and GitHub tools. Additionally, we included some unanswerable questions to gauge the model’s awareness and ability to recognize its limitations.

\section{\textbf{Results and Analysis of Query Handling and Response Generation}}



\subsection{Scope and Limitations Handling}

One of the key aspects of our evaluation was to understand how the GPT-4 model manages queries that fall outside the scope of the available tools. Specifically, we included several questions that required information or functionalities not provided by the current toolset, as a means of testing the model’s ability to recognize its limitations. Our findings indicate that GPT-4 demonstrated a commendable understanding of its scope. When presented with out-of-scope questions, the model did not attempt to fabricate answers; instead, it accurately recognized that the query was beyond its tool’s capabilities and refrained from generating incorrect responses. This behavior is significant, as it highlights the model’s ability to manage user expectations by acknowledging tool constraints rather than attempting to provide inaccurate or misleading information. Such self-awareness in LLMs can improve user trust, as users are less likely to receive potentially erroneous data when tools are unavailable.

\subsection{Iteration Count Analysis}
We analyzed the number of iterations required for each question category. Figure~\ref{fig:bar_plot} summarizes the resulting frequency distribution. This analysis shed light on the efficiency of the model and the work involved in answering different classes of questions. Our findings are as follows:

\begin{figure}[ht]
  \centering
  \includegraphics[width=0.8\columnwidth]{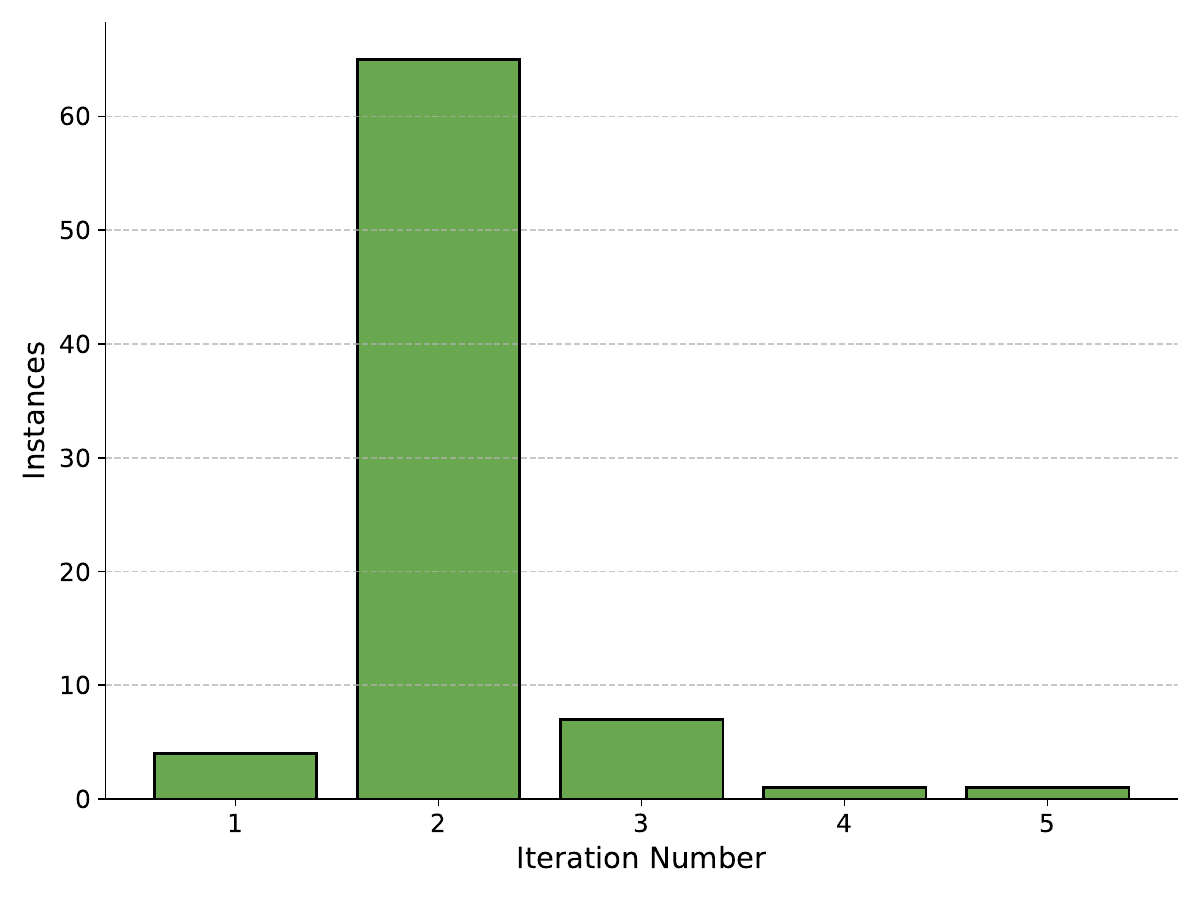}
  \caption{Bar plot showing the frequency distribution of iteration numbers.}
  \label{fig:bar_plot}
\end{figure}

\subsubsection{Simple Queries}
For simple questions like ‘What is the name of the repository?’, the model needed only one iteration, responding quickly without complex processing.

\subsubsection{Two-Iteration Queries}
Most questions required two iterations: the first to determine the appropriate tool (query classification) and the second to generate the final answer. This two-step process was effective for queries closely aligned with the tools’ capabilities.

\subsubsection{Compound Questions}
Compound questions (e.g., “How many issues are open, and who is the repository owner?”) typically required three iterations. The model had to break down, identify, and classify each part, then generate a cohesive response. This approach ensured accurate and complete answers to complex, multi-part queries.

\subsubsection{Error Handling and Out-of-Scope Queries}
In some cases, questions required three iterations for successful data retrieval. If the model struggled to acquire data points initially, a second iteration was needed. When questions were out of scope, iteration counts increased, sometimes up to five, reflecting the model’s attempts to reclassify and process the query before recognizing its limitations.

\subsection{Impact of Tool Availability on Accuracy}

Additionally, we investigated how the availability of tools affected the accuracy of the model's responses. The model currently uses two primary tools: one for generating comprehensive repository reports and another for answering entity-specific questions involving issues, commits, and pull requests. 

However, expanding the toolset could make the system more accurate and versatile. For example, additional tools could support specialized repository-analysis tasks, such as issue-sentiment analysis and contribution-trend analysis. A broader toolset could therefore enable the system to answer a wider range of questions and generate more comprehensive responses. These results indicate that tool availability is an important factor in the effectiveness of an LLM-based repository-analysis chatbot.

\subsection{Evaluation of Model Accuracy}

Each generated response was compared with the corresponding ground-truth answer obtained directly from the GitHub repository data. A response was marked correct only when it returned the expected value and correctly applied all requested filters. An out-of-scope response was marked correct when the system explicitly indicated that the available tools could not answer the query.

We evaluated the model using a dataset of 80 in-scope and out-of-scope questions. Among these, 14 questions were specifically designed to be out of scope. The model correctly identified its limitations and abstained from answering all 14 questions.

For the remaining 66 in-scope questions, the model produced correct answers for 65, achieving an accuracy of 98.48\%. The single incorrect response involved a date-range query in which the model failed to select the correct date parameter and instead relied on the current timestamp. Overall, the model achieved high accuracy on in-scope questions and correctly abstained from answering all out-of-scope questions.

\section{Conclusion}

We introduced a chatbot architecture leveraging OpenAI’s GPT-4 to streamline data extraction and analysis in software repositories. Our solution simplifies the retrieval of repository insights, making it accessible to both technical and non-technical users.

We evaluated our prompts using a custom dataset across Issues, Pull Requests, and Commits, finding that precise prompts enhance model accuracy and efficiency. The tool-based approach allowed GPT-4 to recognize its operational limits and signal out-of-scope queries effectively. Our framework highlights the need for additional specialized tools to improve query coverage, especially for complex questions, potentially increasing accuracy. We demonstrate how targeted prompt engineering and strategic tool use can reduce repository data access complexities and improve collaboration in software development. Future work will extend tool capabilities to support a wider variety of queries.

\bibliography{references}

\end{document}